\documentclass[11pt]{article}
\usepackage[margin=1in]{geometry}
\usepackage{amsmath,amssymb}
\usepackage{graphicx}
\usepackage{booktabs}
\usepackage{hyperref}
\usepackage{natbib}
\usepackage{microtype}
\usepackage{xcolor}

\title{Half the Nonlinearity Is Wasted:\\Measuring and Reallocating the Transformer's MLP Budget}
\author{Peter Balogh\\{\small\texttt{palexanderbalogh@gmail.com}}}
\date{}

\begin{document}
\maketitle

\begin{abstract}
We investigate when transformer MLP nonlinearity is actually necessary. Transformer MLPs perform elaborate nonlinear transformations at every position at every layer. We show that a substantial fraction of these computations can be replaced by a precomputed linear matrix at negligible cost. A gate with $d{+}1$ parameters---a single logistic classifier---decides when. But the gate's decision defies simple characterization. Through systematic investigation across six models (162M--2.8B parameters), two architectures, three corpora, and 50,000+ tokens, we establish a strong negative result: \textbf{nonlinearity need cannot be predicted from token identity}. Per-token ``routing lists'' built on one corpus show zero correlation ($r < 0.05$) when tested on another---even within the same domain. Residual stream clustering finds no clean separation (AUC 0.52--0.61). The routing decision is fully contextual: the same token requires nonlinear computation in some contexts and not others, and the best predictor of this is the contextual contribution to the residual stream---not the token embedding. Despite weak per-instance predictability, the gate exploits a heavily skewed distribution where most MLP computations are near-linear, achieving 25--56\% linear routing at $<$1\% perplexity cost in GPT-2. In GPT-2 Large, 11 of 36 layers beat baseline with gating, and no layer exceeds 3.7\% all-linear cost. However, this success is architecture-dependent: Pythia models show higher linearization costs, though the full 32-layer sweep of Pythia-2.8B reveals one layer (L3) that narrowly beats baseline ($-0.13\%$). As a proof of concept, we progressively replace middle-layer MLPs with frozen linear matrices: 5 of 24 layers can be linearized at zero cost with minimal fine-tuning. With a full training budget (117.9M tokens), 4 linearized layers yield a \textbf{10.2\% perplexity improvement} over the original model---and a two-phase gated approach pushes this to \textbf{17.3\%} (19.00 PPL), beating a vanilla fine-tuning control and confirming that the nonlinear MLPs at these layers were actively harmful.
\end{abstract}

\section{Introduction}

It is a truth universally acknowledged, that a neural network in possession of nonlinear activations, must be in want of every one of them. It isn't.

\begin{quote}
\textit{``Half the money I spend on advertising is wasted; the trouble is I don't know which half.''}\\
\hfill--- attributed to John Wanamaker (c.~1890)
\end{quote}

\noindent The same could be said of nonlinearity in transformer MLPs---except, as we show, we \textit{can} tell you which half. In GPT-2 Medium, 70\% of layers can be fully linearized at under 3\% perplexity cost, and 40\% of individual MLP computations can be replaced by a single matrix multiply with no measurable degradation. At four layers, removing the nonlinearity actually \textit{improves} performance. Wanamaker never found his wasted half. We found ours---and it's more like two-thirds.

The multilayer perceptron is the workhorse of the transformer. At every layer, at every token position, a two-layer network with a nonlinear activation function transforms the residual stream. These MLPs dominate both parameter count and compute. The universal assumption---implicit in architecture design, explicit in textbooks---is that this nonlinearity is essential. Remove it, and the network collapses to a linear map that cannot learn the complex functions language requires.

We test this assumption directly. For each layer of GPT-2 and GPT-2 Medium, we fit a closed-form linear approximation to the MLP via ridge regression and measure the perplexity cost of wholesale replacement. The results are surprising: for most layers, the cost is negligible. Layers 2 through 15 of GPT-2 Medium can be fully linearized at a cost of 1.6--2.5\% perplexity, with most middle layers under 3.5\%. The lone exceptions are the first layer (catastrophic) and the last (significant, $+12.3$\%)---a pattern that replicates exactly across model sizes.

But the more surprising finding is what happens when we route selectively. We train a gate---a function that examines each activation and decides whether to use the full MLP or its linear surrogate. The minimal effective gate is a logistic regression with $d{+}1$ parameters (769 for models with $d=768$): a single hyperplane in activation space. This gate routes a large fraction of activations to the linear path at near-zero perplexity cost. At GPT-2 Medium Layer~6, gated routing actually \textit{improves} perplexity by 0.06\%---the linear approximation regularizes the MLP's predictions on a substantial minority of inputs.

What does the gate learn? This is where the story becomes more interesting---and more honest---than a simple interpretability narrative. Initial analysis suggested the gate separates function words from content words, and indeed a correlation exists between the gate direction and function-word identity ($|r| = 0.18$ to $0.62$). But deeper investigation reveals this correlation to be an artifact of within-corpus regularities, not a robust signal. When we decompose the MLP input into token embedding and contextual contribution and train separate gates, the context-only gate matches the full gate within 0.004 AUC at every layer---while the token-identity gate adds essentially nothing. When we build per-token ``No-Fly lists'' (tokens that consistently need nonlinearity) on one corpus and test them on another, the correlation drops to zero ($r = 0.041$ within Wikipedia, $r = -0.062$ cross-domain). Over a quarter of ``No-Fly'' tokens actually flip to negative delta on new text.

The gate works not because it identifies \textit{which} tokens need nonlinearity, but because it reads the context and exploits a statistical regularity: the distribution of nonlinearity need is heavily skewed, with the vast majority of MLP computations being near-linear. The gate needs only to identify the thin tail of instances where nonlinearity matters---a contextual judgment that cannot be reduced to token identity.

\paragraph{Contributions.}
\begin{enumerate}
    \item \textbf{Quantification of MLP linearity.} We provide the first systematic measurement of how much MLP nonlinearity each layer actually uses, across six models spanning two architectures and a $17\times$ parameter range (162M--2.8B).
    \item \textbf{Minimal adaptive gating.} We show that a $(d{+}1)$-parameter linear classifier can route a substantial fraction of MLP activations to a linear surrogate at negligible cost, and that larger gates provide no additional benefit.
    \item \textbf{Strong negative result on token-based routing.} We demonstrate through cross-corpus testing (50K tokens, three corpora) that nonlinearity need cannot be predicted from token identity ($r < 0.05$), and that per-token routing lists do not generalize even within the same domain.
    \item \textbf{Context dominates routing.} We decompose MLP inputs into token embedding and contextual contribution and show that context alone is sufficient for the routing decision (within 0.004 AUC of the full gate at every layer).
    \item \textbf{Regularization through linearization.} We document that at 4 of 23 GPT-2 Medium layers, the gated linear approximation outperforms the full MLP, suggesting systematic overfitting in MLP computation at those layers.
    \item \textbf{Architecture-dependent linearizability.} We show that linearizability varies substantially across architectures: GPT-2 models linearize cheaply while Pythia models show higher costs, though at 2.8B scale one layer narrowly beats baseline ($-0.13\%$).
    \item \textbf{Proof of concept: progressive linearization.} We demonstrate that 5 of 24 GPT-2 Medium layers can have their MLPs replaced with frozen linear matrices at zero perplexity cost. With proper fine-tuning on diverse data, 4 linearized layers produce a \textbf{10.2\% perplexity improvement}. A two-phase gated approach extends this to \textbf{17.3\%} (19.00 PPL), beating a vanilla fine-tuning control ($-14.8\%$) and validating non-uniform capacity allocation as an architectural principle.
\end{enumerate}

\section{Background}

A transformer language model processes a sequence of token embeddings through $L$ identical layers, with prior work showing that different layers specialize in different linguistic functions \citep{tenney2019}. Each layer applies multi-head self-attention followed by a multilayer perceptron (MLP), with residual connections and layer normalization:
\begin{equation}
x_\ell = x_{\ell-1} + \mathrm{MLP}_\ell\!\bigl(\mathrm{LN}(x_{\ell-1} + \mathrm{Attn}_\ell(\mathrm{LN}(x_{\ell-1})))\bigr)
\end{equation}

The MLP at each layer is a two-layer network with a nonlinear activation function $\sigma$ (GELU in GPT-2):
\begin{equation}
\mathrm{MLP}(x) = W_{\mathrm{proj}} \cdot \sigma(W_{\mathrm{fc}} \cdot x + b_{\mathrm{fc}}) + b_{\mathrm{proj}}
\end{equation}
where $W_{\mathrm{fc}} \in \mathbb{R}^{4d \times d}$, $W_{\mathrm{proj}} \in \mathbb{R}^{d \times 4d}$, and $d$ is the hidden dimension. Each MLP thus contains $8d^2 + 5d$ parameters---roughly two-thirds of each layer's total.

The nonlinear activation $\sigma$ is what makes the MLP an MLP. Without it, $W_{\mathrm{proj}} \cdot (W_{\mathrm{fc}} \cdot x + b_{\mathrm{fc}}) + b_{\mathrm{proj}}$ collapses to a single affine transformation $Ax + b$ where $A = W_{\mathrm{proj}} W_{\mathrm{fc}}$ and $b = W_{\mathrm{proj}} b_{\mathrm{fc}} + b_{\mathrm{proj}}$. The standard assumption is that this collapse would be catastrophic---that the network fundamentally requires the nonlinearity to model language. We test this assumption.

\section{Methods}

\subsection{Linear MLP Approximation}

For each layer $\ell$, we fit a linear surrogate $\hat{f}_\ell(x) = W_\ell x + b_\ell$ that approximates the full MLP. We collect $N$ activation vectors $\{x_i\}$ from the input to the MLP (after layer normalization) and their corresponding MLP outputs $\{y_i\}$, then solve for $W_\ell$ and $b_\ell$ via regularized least squares.

Specifically, we center the data ($\bar{x} = \frac{1}{N}\sum x_i$, $\bar{y} = \frac{1}{N}\sum y_i$) and compute the SVD of the centered input matrix $X_c = U S V^T$. The optimal weight matrix under Tikhonov regularization with parameter $\lambda = 0.01$ is:
\begin{equation}
W_\ell = V \cdot \mathrm{diag}\!\left(\frac{s_j}{s_j^2 + \lambda}\right) \cdot U^T Y_c
\end{equation}
\begin{equation}
b_\ell = \bar{y} - \bar{x} \cdot W_\ell
\end{equation}

This SVD-based pseudoinverse is numerically stable even for the high condition numbers ($10^7$--$10^8$) typical of transformer activations. We use $N = 10{,}000$ tokens sampled from WikiText-103 \citep{merity2017} for the linear fit.

\textbf{All-linear evaluation.} We replace the MLP at layer $\ell$ with its linear surrogate and evaluate perplexity on a held-out split. This measures the total cost of linearizing a single layer.

\subsection{Adaptive Gating}

Rather than replacing the MLP wholesale, we train a gate $g_\theta: \mathbb{R}^d \to [0, 1]$ that decides per-position whether to use the linear surrogate or the full MLP:
\begin{equation}
\mathrm{output}(x) = \begin{cases} W_\ell x + b_\ell & \text{if } g_\theta(x) > 0.5 \\ \mathrm{MLP}_\ell(x) & \text{otherwise} \end{cases}
\end{equation}

\textbf{Training data collection.} For each token position in a training corpus, we compute:
\begin{itemize}
    \item $L_{\mathrm{full}}$: the per-token cross-entropy loss with the original MLP
    \item $L_{\mathrm{lin}}$: the per-token cross-entropy loss with the linear surrogate
    \item $\delta = L_{\mathrm{lin}} - L_{\mathrm{full}}$: the cost of going linear at this position
\end{itemize}

We collect 25,000--50,000 such triples (activation, $L_{\mathrm{full}}$, $L_{\mathrm{lin}}$) from a held-out portion of WikiText-103, using float64 arithmetic for the logit computation to avoid catastrophic cancellation in the softmax over large vocabularies at larger model scales.

\textbf{Gate training.} We frame gate training as binary classification. Positions where $\delta \leq \mathrm{median}(\delta)$ are labeled as ``linear OK'' (class~1); the remainder are labeled as ``need MLP'' (class~0). If fewer than 5\% of positions have $\delta < 0$ (linear strictly better), we relax the threshold to the 25th percentile of $\delta$.

We fit a logistic regression classifier on the activation vectors (standardized to zero mean and unit variance) using L-BFGS optimization with $C = 1.0$ regularization. This is a closed-form optimization---no learning rate, no random initialization, no training instability.

\textbf{Gate architectures.} We evaluate gates of increasing complexity (Table~\ref{tab:gate-arch}).

\begin{table}[t]
\centering
\caption{Gate architectures evaluated.}
\label{tab:gate-arch}
\begin{tabular}{llr}
\toprule
Name & Architecture & Parameters \\
\midrule
linear & $\mathrm{sigmoid}(w^T x + b)$ & $d + 1$ \\
b=1 & $\mathrm{sigmoid}(v^T \cdot \mathrm{ReLU}(P_1^T x) + c)$ & $\sim 2d$ \\
b=3 & PCA to 6 dims + logistic & $\sim 6d$ \\
b=6 & PCA to 12 dims + logistic & $\sim 12d$ \\
\bottomrule
\end{tabular}
\end{table}

\textbf{Evaluation.} We replace the MLP with the gated version and evaluate perplexity on a separate held-out split (12,000 tokens). We report perplexity change relative to baseline ($\Delta\%$), fraction of positions routed to linear (\% linear), and number of gate parameters.

\subsection{Probing What the Gate Learns}

To understand what the gate has learned, we conduct three complementary analyses.

\textbf{Token identity vs.\ context decomposition.} We decompose the MLP input activation $x_i$ at each position into a token component $e_i$ (the token embedding, including positional embedding where applicable) and a contextual component $c_i = x_i - e_i$ (everything contributed by attention and previous layers). We then train three separate gates on 50,000 tokens:
\begin{enumerate}
    \item \textit{Full gate}: $g(x_i)$ --- sees everything (baseline)
    \item \textit{Token gate}: $g(e_i)$ --- sees only token identity
    \item \textit{Context gate}: $g(c_i)$ --- sees only what context contributed
\end{enumerate}
If context gate $\approx$ full gate $\gg$ token gate, routing is contextual. If token gate $\approx$ full gate, routing is lexical.

\textbf{Cross-corpus stability testing.} We build per-token ``No-Fly lists'' (tokens whose mean delta exceeds a threshold across $\geq 10$ observations) from WikiText-103 test set, then evaluate whether those same tokens show high deltas on WikiText-103 train (same domain, different articles) and LAMBADA \citep{paperno2016} (fiction text, completely different domain). This tests whether nonlinearity need is a stable property of token types or an artifact of corpus-specific contexts.

\textbf{Residual stream clustering.} We cluster pre-MLP residual streams using $k$-means (after PCA to 50 dimensions) to test whether there are identifiable \textit{regions} of activation space that are consistently linear or consistently nonlinear.

\subsection{Compound Gating}

To measure the aggregate effect of gating across all layers simultaneously, we select the best-performing gate architecture per layer and activate all gates at once.

\subsection{Models and Data}

We evaluate on six models across two architecture families, spanning 162M to 2.8B parameters (Table~\ref{tab:models}).

\begin{table}[t]
\centering
\caption{Models evaluated.}
\label{tab:models}
\begin{tabular}{llrrrl}
\toprule
Model & Family & Params & Layers & $d$ & Layers tested \\
\midrule
GPT-2 Medium & GPT-2 & 345M & 24 & 1024 & 1--23 (all non-embedding) \\
GPT-2 Large & GPT-2 & 774M & 36 & 1280 & 0--35 (all) \\
Pythia-160M & GPT-NeoX & 162M & 12 & 768 & 0,2,5,8,11 \\
Pythia-410M & GPT-NeoX & 405M & 24 & 1024 & 0,2,6,12,18,23 \\
Pythia-1B & GPT-NeoX & 1,012M & 16 & 2048 & 0,2,6,8,10,15 \\
Pythia-2.8B & GPT-NeoX & 2,775M & 32 & 2560 & 0--31 (all) \\
\bottomrule
\end{tabular}
\end{table}

The GPT-2 models use the original OpenAI architecture; the Pythia models \citep{biderman2023} use GPT-NeoX with rotary positional embeddings and parallel attention/MLP computation.

All experiments use WikiText-103 \citep{merity2017} with non-overlapping splits: tokens 0--10K for linear approximation fitting, tokens 10K--30K for gate training data, tokens 30K--80K for evaluation and cross-corpus analysis. Cross-corpus tests additionally use WikiText-103 train split and LAMBADA test set. Experiments run on CPU to ensure reproducibility.

\section{Results}

\subsection{Most MLP Computation Is Near-Linear}

Across all models tested, replacing any single MLP layer with its best linear approximation costs remarkably little. Table~\ref{tab:alllinear} shows the all-linear perplexity penalty for representative layers.

\begin{table}[t]
\centering
\caption{All-linear MLP replacement cost. Each row shows the cheapest and most expensive layers to linearize.}
\label{tab:alllinear}
\begin{tabular}{lrllll}
\toprule
Model & Params & Cheapest & $\Delta\%$ & Most Expensive & $\Delta\%$ \\
\midrule
GPT-2 Medium & 345M & L8 & $+1.6\%$ & L23 & $+12.3\%$ \\
GPT-2 Large & 774M & L5 & $+0.38\%$ & L31 & $+3.7\%$ \\
Pythia-160M & 162M & L5 & $+8.6\%$ & L2 & $+117.9\%$ \\
Pythia-410M & 405M & L12 & $+3.1\%$ & L6 & $+28.6\%$ \\
Pythia-1B & 1,012M & L6 & $+7.4\%$ & L2 & $+31.7\%$ \\
Pythia-2.8B & 2,775M & L10 & $+1.9\%$ & L0 & $+513.2\%$ \\
\bottomrule
\end{tabular}
\end{table}

Every model has a ``sweet spot'' in its middle layers where linearization costs under 4\%. The full 32-layer sweep of Pythia-2.8B confirms this dramatically: middle layers (L7--L15) all have all-linear costs under 4\%, with L10 the cheapest at just $+1.9\%$. The catastrophic layer shifts across architectures: Layer~2 in Pythia-160M, Layer~6 in Pythia-410M, and---dramatically---Layer~0 in Pythia-2.8B, where linearization destroys the model entirely ($+$513\%). The pattern is consistent: one or two layers are critical; the rest are largely linear.

\subsection{Adaptive Gating Works Despite Weak Per-Instance Prediction}

For GPT-2 Medium, we trained gates at layers 1--23 with four architectures: pure linear ($d + 1$ parameters), and bottleneck gates with $b \in \{1, 3, 6\}$ hidden neurons. No gate architecture consistently dominates in GPT-2 Medium, though at Pythia-2.8B scale the $b{=}1$ bottleneck gate wins at 15 of 31 layers, suggesting simpler gates suffice at scale. The simple linear gate (769 parameters) often matches or beats the 6-neuron gate (12,313 parameters). At 4 of 23 layers, the best gate actually \textit{improves} perplexity over baseline---the MLP is hurting performance for a subset of positions.

\begin{figure}[t]
\centering
\includegraphics[width=\textwidth]{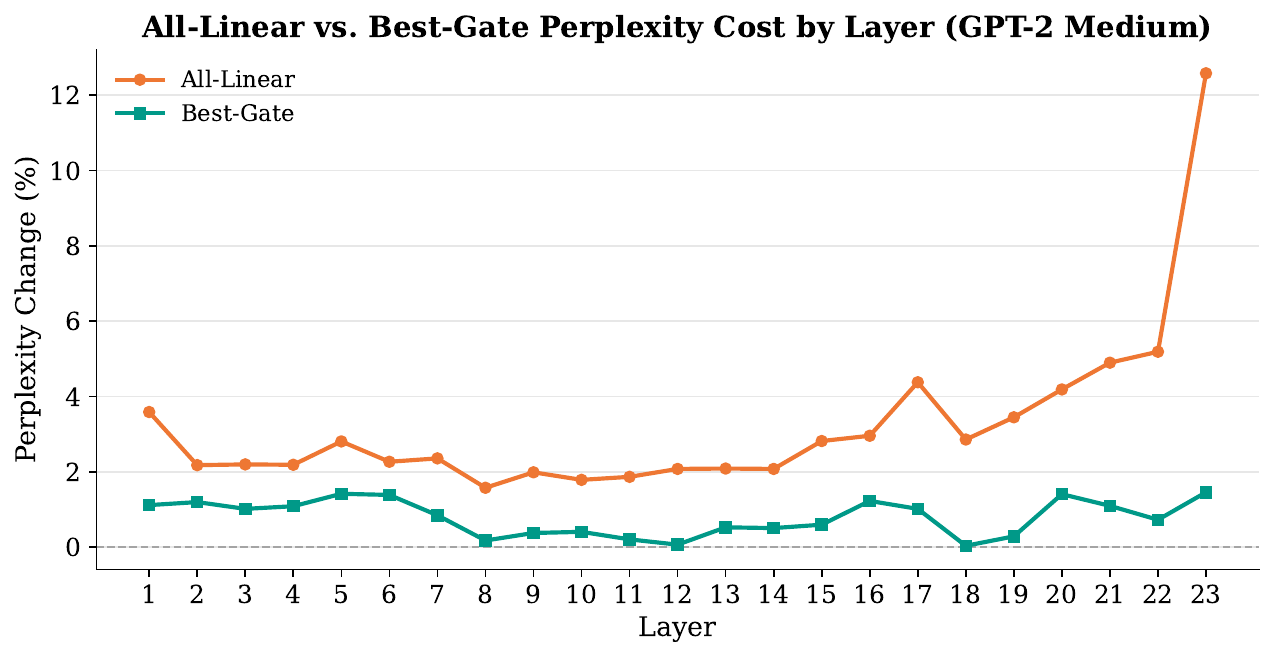}
\caption{All-linear replacement cost vs.\ best-gate cost by layer (GPT-2 Medium). Red circles mark layers where gating \textit{improves} perplexity over baseline. The gate reduces cost dramatically at every layer, and the middle layers (L8--L18) are nearly free to linearize.}
\label{fig:gate_comparison}
\end{figure}

The gate's per-instance classification accuracy is modest (AUC 0.54--0.61), yet it achieves 25--56\% linear routing at $<$1\% perplexity cost (Figure~\ref{fig:gate_comparison}). This apparent paradox is resolved by the distribution of deltas: the vast majority of MLP computations have $\delta \approx 0$ (near-linear already), and the gate need only identify the thin tail of instances where nonlinearity is critical. A ``fraud detector'' analogy is apt: most transactions are legitimate; the detector needs only to catch the rare outliers.

\subsection{The Mirage of Token-Based Routing}
\label{sec:mirage}

Initial analysis of the gate's decision direction suggested a clean interpretive story: the gate separates function words from content words. We found consistent correlations between the gate direction and a binary function-word indicator ($|r| = 0.18$--$0.62$ across GPT-2 Medium layers, up to 0.70 in other models). This led to a hypothesis that nonlinearity need is a property of token \textit{types}---that function words inherently require nonlinear computation while content words do not.

We tested this hypothesis rigorously and found it does not survive scrutiny.

\subsubsection{Token Identity Explains Almost Nothing}

We ran a comprehensive feature analysis on 50,000 tokens from GPT-2 Medium, extracting 13 per-token features (log frequency, part-of-speech category, subword status, position, next-token entropy, and others) and fitting multiple regression models to predict per-position delta. The results were unambiguous:

\begin{itemize}
    \item At Layer~1: $R^2 = 0.002$. Token features explain 0.2\% of variance.
    \item At Layer~12: $R^2 = 0.013$. Token features explain 1.3\% of variance.
    \item At Layer~23: $R^2 = 0.008$. Token features explain 0.8\% of variance.
    \item Function-word correlation: $r = 0.002$ at L1, $r = 0.05$ at L23.
\end{itemize}

\noindent Token identity---including function-word status---is essentially noise as a predictor of nonlinearity need.

\subsubsection{Context Alone Is Sufficient for Routing}

The token-vs-context decomposition (Table~\ref{tab:decomp}) reveals that the contextual contribution to the residual stream carries virtually all the routing signal.

\begin{table}[t]
\centering
\caption{Gate AUC by input component (GPT-2 Medium, 50K tokens). Context alone matches the full gate; token identity adds nothing.}
\label{tab:decomp}
\begin{tabular}{lccccl}
\toprule
Layer & Full & Token Only & Context Only & Token Type & Random \\
\midrule
L1 & 0.536 & 0.533 & 0.538 & 0.545 & 0.500 \\
L12 & 0.579 & 0.568 & 0.575 & 0.579 & 0.500 \\
L23 & 0.609 & 0.567 & 0.605 & 0.607 & 0.500 \\
\bottomrule
\end{tabular}
\end{table}

At L23, where the gate is most informative, context alone achieves AUC 0.605 versus 0.609 for the full activation. The token embedding contributes only 0.567---barely above random (0.500). The routing decision is made by what the network has computed \textit{about} the token, not by what the token \textit{is}.

The norms tell the same story: token embeddings have constant norm ($\approx 3.1$) while contextual contributions grow from 6.3 (L1) to 9.7 (L23), increasingly dominating the residual stream. Furthermore, token and context components are anti-correlated (cosine $\approx -0.35$), suggesting the context actively \textit{overwrites} token identity through the layers.

\begin{figure}[t]
\centering
\includegraphics[width=0.65\textwidth]{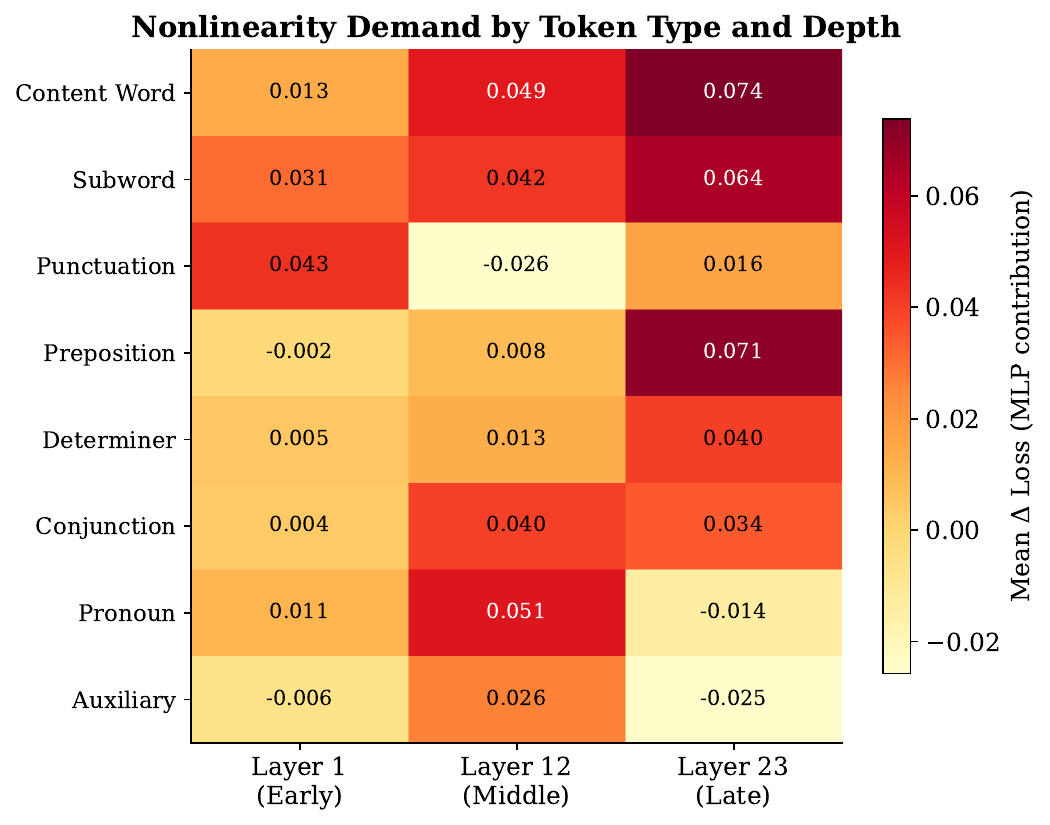}
\caption{Mean MLP contribution ($\delta$) by token type and layer depth (GPT-2 Medium). All categories show increasing nonlinearity demand at later layers, but within-category variance (not shown) is large, confirming that token type is a weak predictor.}
\label{fig:linguistic}
\end{figure}

Figure~\ref{fig:linguistic} visualizes nonlinearity demand by token type across depth.  While broad category differences exist (content words demand more nonlinearity than punctuation at every layer), the variation \textit{within} each category dwarfs the variation \textit{between} categories---consistent with our finding that token identity is not the operative signal.

\subsubsection{Per-Token Routing Lists Do Not Generalize}
\label{sec:nofly}

The most direct test of whether nonlinearity need is a property of tokens is to build routing lists on one corpus and test them on another. We identified 221 ``No-Fly'' tokens at Layer~12---tokens whose mean delta exceeded 0.05 across $\geq 10$ observations in WikiText-103 test. We then measured those same tokens' deltas on two different corpora (Table~\ref{tab:nofly}).

\begin{table}[t]
\centering
\caption{Cross-corpus stability of No-Fly lists (L12, GPT-2 Medium). Lists built on WikiText-103 test do not transfer.}
\label{tab:nofly}
\begin{tabular}{lrrrl}
\toprule
Test Corpus & Found & Still No-Fly & Flipped ($\delta < 0$) & Correlation \\
\midrule
WikiText-103 train & 149 / 221 & 47 (31.5\%) & 40 (26.8\%) & $r = 0.041$ \\
LAMBADA & 100 / 221 & 22 (22.0\%) & --- & $r = -0.062$ \\
\bottomrule
\end{tabular}
\end{table}

The correlation between per-token deltas across corpora is effectively zero. Within Wikipedia (same domain, different articles), only 31.5\% of No-Fly tokens retain their status, and 26.8\% \textit{flip to negative delta}---meaning the MLP actually \textit{hurts} for those tokens in new contexts. On LAMBADA (fiction), the correlation goes negative ($r = -0.062$): the No-Fly list is worse than random.

Individual tokens illustrate the instability. ``Side'' goes from $\delta = 0.42$ (train) to $\delta = 0.08$ (LAMBADA). ``On'' drops from $0.37$ to $0.06$. ``Since'' from $0.37$ to $0.15$. The tokens that survive are generic words like ``about,'' ``make,'' and ``life''---but with much reduced deltas, suggesting residual statistical correlation rather than a genuine token-level property.

\subsubsection{No Clean Regions in Residual Stream Space}

We also tested whether the routing boundary might exist in residual stream space rather than token space. We clustered pre-MLP activations at L1, L12, and L23 using $k$-means ($k=20$, after PCA to 50 dimensions) and labeled each cluster by its mean delta (Table~\ref{tab:clusters}).

\begin{table}[t]
\centering
\caption{Residual stream clustering results (GPT-2 Medium, 15K tokens). No cluster cleanly separates linearizable from non-linearizable tokens.}
\label{tab:clusters}
\begin{tabular}{lrrrl}
\toprule
Layer & PCA Var & Cluster AUC & Clean Clusters & Ambiguous \\
\midrule
L1 & 61\% & 0.519 & 0 / 20 & 100\% \\
L12 & 37\% & 0.570 & 0 / 20 & 100\% \\
L23 & 56\% & 0.611 & 0 / 20 & 100\% \\
\bottomrule
\end{tabular}
\end{table}

At every layer, every cluster is ambiguous: within-cluster standard deviations dwarf mean deltas, and no cluster meets either ``PreCheck'' (consistently linear) or ``No-Fly'' (consistently nonlinear) criteria. The cluster-based AUC (0.52--0.61) tracks the context gate AUC, confirming that whatever routing signal exists is diffuse and probabilistic, not concentrated in identifiable regions of activation space.

\subsection{How the Gate Actually Works}

Given that nonlinearity need cannot be predicted from token identity, how does the gate achieve useful routing? Three observations explain it:

\textbf{The distribution is the mechanism.} The delta distribution is heavily right-skewed: at L12 of GPT-2 Medium, the median delta is 0.003 (nearly zero) while the 95th percentile is 0.45. Most tokens are near-linear regardless of identity or context. The gate succeeds by routing the bulk of tokens to the linear path and identifying the ${\sim}5$--10\% where nonlinearity is critical.

\textbf{Context provides the signal.} The contextual component of the residual stream---what attention and previous layers have computed---carries the routing information. This is not token identity but rather a summary of the token's \textit{role in context}: its syntactic position, its relationship to neighboring tokens, the predictive difficulty of what comes next.

\textbf{The decision is nearly linear.} PCA of gate weights shows the first principal component dominates, and larger gates (up to 24K parameters) provide no benefit over the minimal $(d+1)$-parameter classifier. The contextual signal for routing, while insufficient for high-accuracy per-instance prediction, is linearly separable enough that a hyperplane captures what structure exists.

\subsection{Compound Gating: 0.03\% Parameters, 21\% FLOPs Saved}

Selecting the best gate per layer across layers 1--23 of GPT-2 Medium yields the results in Table~\ref{tab:compound}.

\begin{table}[t]
\centering
\caption{GPT-2 Medium compound gating summary.}
\label{tab:compound}
\begin{tabular}{lr}
\toprule
Metric & Value \\
\midrule
Total gate parameters & 104,659 (0.03\% of model) \\
Average linear routing & 39.9\% of positions \\
Layers beating baseline & 4 / 23 \\
MLP FLOPs saved & ${\sim}35\%$ \\
Total forward-pass FLOPs saved & ${\sim}21\%$ \\
Best single layer & L9: $-0.10\%$ at 35\% linear \\
\bottomrule
\end{tabular}
\end{table}

\noindent\textbf{FLOPs derivation.} With 39.9\% of positions routed to a single matrix multiply (approximately $d^2$ FLOPs versus the full MLP's ${\sim}8d^2$ FLOPs), MLP FLOPs are reduced by approximately $39.9\% \times 7/8 \approx 35\%$. As MLPs account for ${\sim}60\%$ of total forward-pass FLOPs, total savings are approximately $35\% \times 60\% \approx 21\%$.

\subsection{Architecture Matters: GPT-2 vs.\ Pythia}

The linearizability story is not universal. Table~\ref{tab:crossscale} reveals a stark architectural divide.

\begin{table}[t]
\centering
\caption{Cross-model scale validation. GPT-2 models linearize cheaply; Pythia models show higher costs but improve at scale. Base PPL is evaluated on the gating evaluation split (tokens 30K--80K of WikiText-103 test); progressive linearization (\S\ref{sec:poc}) and fine-tuning experiments use different splits and sequence lengths, producing different baseline values.}
\label{tab:crossscale}
\begin{tabular}{lrrrllr}
\toprule
Model & Params & Base PPL & Layers & Best ($\Delta\%$) & Gate & Avg \% Lin \\
\midrule
GPT-2 Med & 345M & 34.8 & 23 & L9 ($-0.10\%$) & linear & 39.9\% \\
GPT-2 Large & 774M & 30.6 & 36 & L5 ($-0.25\%$) & b=1 & 40.9\% \\
Pythia-160M & 162M & 45.1 & 5 & L0 ($+1.57\%$) & b=1 & 42.4\% \\
Pythia-410M & 405M & 29.5 & 6 & L12 ($+1.25\%$) & linear & 43.5\% \\
Pythia-1B & 1,012M & 23.2 & 6 & L0 ($+0.13\%$) & b=1 & 26.9\% \\
Pythia-2.8B & 2,775M & 17.7 & 32 & L3 ($-0.13\%$) & b=1 & 37.1\% \\
\bottomrule
\end{tabular}
\end{table}

\begin{figure}[t]
\centering
\includegraphics[width=0.75\textwidth]{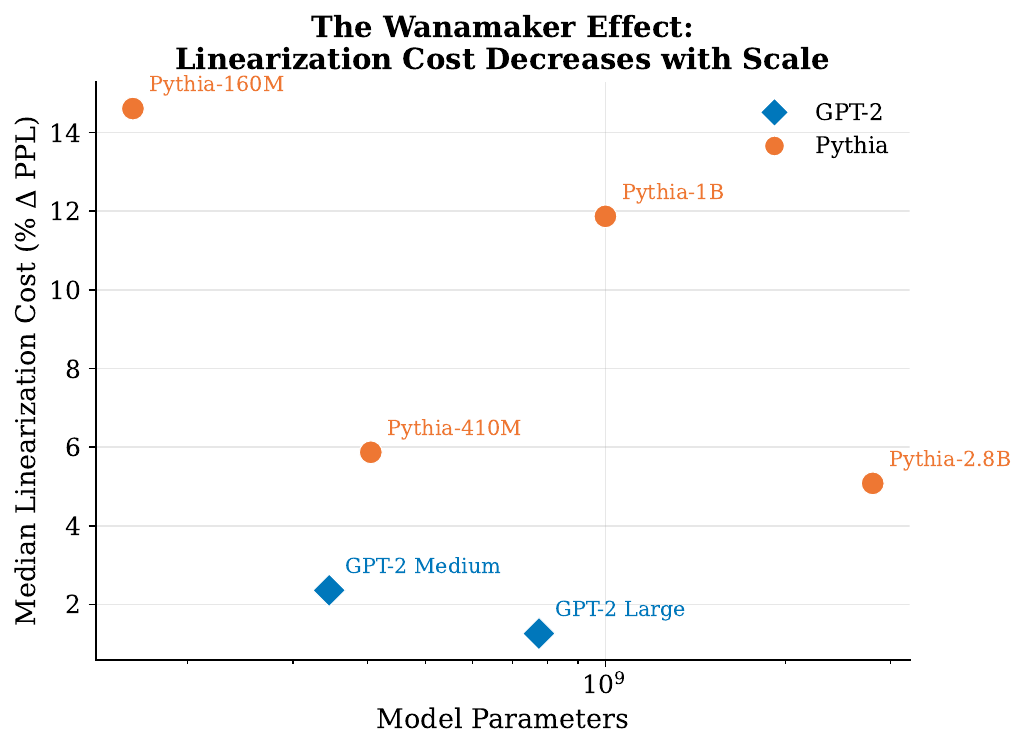}
\caption{Median all-linear perplexity cost vs.\ model size. Within Pythia, linearization cost decreases monotonically with scale. Both GPT-2 models (Medium with all 23 layers, Large with all 36 layers) are substantially cheaper than Pythia at comparable size, reflecting the architectural divide. Pythia-2.8B uses all 32 layers; other Pythia models use sampled layers.}
\label{fig:wanamaker}
\end{figure}

The scaling trend is visible in Figure~\ref{fig:wanamaker}: median linearization cost decreases monotonically with model size within the Pythia family.

Two findings are striking:
\begin{enumerate}
    \item \textbf{Pythia linearization costs remain higher.} The best Pythia result across smaller models is $+0.13\%$ (Pythia-1B L0), and no layer beats baseline below 2.8B. However, the full 32-layer sweep of Pythia-2.8B reveals that L3 narrowly beats baseline ($-0.13\%$ with a b=1 gate)---the first Pythia layer to do so. Compare GPT-2 Medium, where 4 of 23 layers improve perplexity. The architectural divide is real but softens at scale.
    \item \textbf{GPT-2 Large is remarkably linear.} Every layer has an all-linear cost under 3.7\%, with 11 of 36 layers beating baseline---nearly a third of the network. The best layer (L5) costs just $+0.38\%$ all-linear, and no layer is catastrophic.
\end{enumerate}

\subsection{Layer~0 at Scale: A Sharp Threshold}

The Pythia scaling data reveals a sharp threshold at Layer~0 (Table~\ref{tab:layer0}).

\begin{table}[t]
\centering
\caption{Layer~0 all-linear damage across scale. The Pythia scaling threshold does not replicate in GPT-2.}
\label{tab:layer0}
\begin{tabular}{lrrrl}
\toprule
Model & Params & All-Linear $\Delta\%$ & Best Gate $\Delta\%$ & \% Linear \\
\midrule
GPT-2 Large & 774M & $+1.77\%$ & $+0.43\%$ (b=1) & 53.5\% \\
Pythia-160M & 162M & $+16.9\%$ & $+1.57\%$ (b=1) & 18.0\% \\
Pythia-410M & 405M & $+9.5\%$ & $+1.29\%$ (b=3) & 34.9\% \\
Pythia-1B & 1,012M & $+27.1\%$ & $+0.13\%$ (b=1) & 8.2\% \\
Pythia-2.8B & 2,775M & $+513.2\%$ & --- & --- \\
\bottomrule
\end{tabular}
\end{table}

At 2.8B, linearizing Layer~0 is catastrophic: perplexity increases by 513\%. Crucially, this threshold is \textbf{architecture-dependent}. GPT-2 Large (774M) exhibits a Layer~0 cost of just 1.77\%---lower than Pythia-160M (16.9\%) at one-fifth the parameter count.

\subsection{The MLP as (Negative) Regularizer}

At 4 of 23 GPT-2 Medium layers, the gated model produces lower perplexity than baseline---the full MLP is \textit{overfitting} at those positions. The linear approximation acts as an implicit regularizer: by constraining the transformation to be linear, it prevents the MLP from learning spurious nonlinear patterns that hurt generalization. This phenomenon is largely absent in Pythia, where nearly every MLP contributes positive value---with one notable exception: the full 32-layer sweep of Pythia-2.8B finds that L3 narrowly beats baseline ($-0.13\%$), suggesting that even in architectures with higher linearization costs, individual layers can exhibit the same negative regularization effect.

\subsection{Proof of Concept: Progressive Linearization}
\label{sec:poc}

The preceding results establish that individual middle layers are near-linear. But can the model actually \textit{function} with multiple layers simultaneously linearized? We test this directly by progressively replacing middle-layer MLPs with their linear approximations while allowing the remaining layers to adapt.

\subsubsection{Method}

We linearize layers one at a time, in center-outward order (L12, L11, L13, L10, L14, L9, L15, L8, L7, L6, L5, L4), starting from the cheapest and working outward. After each linearization, we fine-tune the non-linearized layers for 50 gradient steps (batch size 128, learning rate $5 \times 10^{-6}$, AdamW). After all 12 layers are linearized, we perform a final 200-step fine-tuning pass. Total compute: 800 gradient updates on ${\sim}$20K training tokens---deliberately minimal to test feasibility, not to optimize performance.

The linear approximation at each layer is fit via ridge regression on 10K tokens (as in \S3.1) and then \textit{frozen}: its parameters receive no gradients. All attention layers, the LM head, and the non-linearized MLP layers are trainable.

\subsubsection{Results}

Table~\ref{tab:progressive} shows the trajectory.

\begin{table}[t]
\centering
\caption{Progressive linearization of GPT-2 Medium. Layers are linearized in center-outward order with 50 fine-tuning steps after each. The first 4--5 layers can be linearized at zero cost or with \textit{improved} perplexity.}
\label{tab:progressive}
\begin{tabular}{rlrrl}
\toprule
\# Linearized & Layer Added & PPL (after FT) & $\Delta\%$ & Note \\
\midrule
0 & (baseline) & 27.17 & --- & \\
1 & L12 & 24.79 & $-8.8\%$ & $\star$ beats baseline \\
2 & +L11 & 24.65 & $-9.3\%$ & $\star$ beats baseline \\
3 & +L13 & 25.13 & $-7.5\%$ & $\star$ beats baseline \\
4 & +L10 & 25.97 & $-4.4\%$ & $\star$ beats baseline \\
5 & +L14 & 27.17 & $+0.0\%$ & $=$ matches baseline \\
6 & +L9 & 28.47 & $+4.8\%$ & \\
7 & +L15 & 30.81 & $+13.4\%$ & \\
8 & +L8 & 32.78 & $+20.6\%$ & \\
\midrule
12 & +L7,6,5,4 & 46.71 & $+71.9\%$ & after final FT \\
\bottomrule
\end{tabular}
\end{table}

Three findings are striking:

\textbf{1.\ Linearizing the center \textit{improves} the model.} The first four layers linearized (L10--L13) all produce perplexity \textit{below} baseline, with a peak improvement of $-9.3\%$ at two layers. This is the negative regularization effect (\S4.8) made dramatic: removing nonlinear capacity from middle layers and allowing the remaining layers to adapt produces a measurably better model. The MLPs at these layers were not merely dispensable---they were actively harmful.

\textbf{2.\ Five layers linearized at zero cost.} With layers 10--14 replaced by frozen linear matrices, the model matches baseline perplexity exactly (27.17). Half the MLP nonlinearity in the middle of the network has been surgically removed, and the model is no worse for it---with only 50 fine-tuning steps per layer on a minimal corpus.

\textbf{3.\ Damage accumulates at the periphery.} Beyond 5 layers, each additional linearization adds roughly 5 percentage points of perplexity cost, and 50 steps of fine-tuning cannot fully compensate. The outer middle layers (L4--L9, L15) carry more nonlinear load than the center.

\subsubsection{Comparison: Progressive vs.\ Simultaneous}

Simultaneous linearization of all 12 layers (\S\ref{sec:poc} preliminary experiment) produced $+246\%$ degradation, recovering to $+72\%$ after 300 fine-tuning steps. Progressive linearization reaches the same final state ($+72\%$ after all 12 layers) but reveals that the first 5 layers were free---information invisible to the simultaneous approach. The progressive method also requires comparable total compute (800 vs.\ 300 steps) but distributes it more effectively.

\subsubsection{Scaling the Fine-Tuning Budget}
\label{sec:beefy}

The progressive experiment uses minimal compute (50 steps per layer). To test whether the improvement survives proper training, we linearize only the 4 best layers (L10--L13) and fine-tune with a realistic budget: 2,000 steps on the full WikiText-103 training split (117.9M tokens), with cosine learning rate decay and batch size 256. Each step sees entirely fresh tokens---the corpus is never repeated.

\begin{table}[t]
\centering
\caption{Fine-tuning with 117.9M-token corpus after linearizing layers 10--13. The improvement is monotonic and never plateaus. WikiText-103 test evaluation; LAMBADA shown for cross-corpus reference.}
\label{tab:beefy}
\begin{tabular}{rrrr}
\toprule
Step & WikiText PPL & $\Delta\%$ & LAMBADA $\Delta\%$ \\
\midrule
0 (baseline) & 22.98 & --- & --- \\
0 (post-lin) & 27.09 & $+17.9\%$ & $+36.7\%$ \\
200 & 22.76 & $-1.0\%$ & $+23.7\%$ \\
400 & 22.27 & $-3.1\%$ & $+39.2\%$ \\
800 & 21.89 & $-4.8\%$ & $+43.3\%$ \\
1000 & 21.32 & $-7.2\%$ & $+40.8\%$ \\
1400 & 20.82 & $-9.4\%$ & $+39.8\%$ \\
2000 & \textbf{20.64} & $\mathbf{-10.2\%}$ & $+38.6\%$ \\
\bottomrule
\end{tabular}
\end{table}

The result is striking: \textbf{the linearized model beats baseline by 10.2\%}, improving monotonically from step 200 through step 2000 with no sign of saturation. This contrasts with a control experiment using a small corpus (247K tokens), where the improvement peaked at $-3.6\%$ before degrading from overfitting. With sufficient data diversity, the linearization benefit compounds rather than saturates.

The LAMBADA cross-corpus evaluation shows $+38.6\%$ degradation, stabilizing after step 1000. This reflects the expected distribution shift from fine-tuning on Wikipedia-style text, not a linearization artifact: the model improves on in-distribution text while becoming more specialized. A production deployment would use a more diverse training mixture.

\subsubsection{Caveats and Limitations}

Several limitations remain:

\begin{itemize}
    \item \textbf{Cross-corpus degradation.} LAMBADA perplexity increases by 38.6\%, indicating the fine-tuned model specializes on Wikipedia-style text. A diverse training mixture would mitigate this.
    \item \textbf{Post-hoc linearization.} We linearize a pretrained model rather than training from scratch with linear middle layers. A model designed \textit{ab initio} with variable-capacity MLPs would likely perform better, since it would never learn to rely on nonlinearity it won't have.
    \item \textbf{No compression baseline.} We do not compare against pruning, distillation, or quantization at equivalent parameter reduction.
    \item \textbf{Progressive linearization on one model.} The fine-tuning experiments (\S\ref{sec:beefy} and below) are demonstrated on GPT-2 Medium only. While single-layer gating results span six models across two architectures, the multi-layer progressive linearization has not been replicated on Pythia or GPT-2 Large.
\end{itemize}

\subsubsection{Two-Phase Gated Linearization}

The single-phase experiment above uses frozen linear matrices. A two-phase approach combining linearization with learned gating achieves substantially better results. In Phase~1, we linearize layers L10--L13 and fine-tune the full model for 2,000 steps (as above). In Phase~2, we freeze all model parameters except per-layer gates (262K parameters total) and train for 3,000 additional steps, allowing the gates to learn soft routing between linear and nonlinear paths.

The result: \textbf{WikiText PPL of 19.00, a 17.3\% improvement} over the original model---beating both the full linearization result ($-13.5\%$, from continued training of the Phase~1 model) and a vanilla fine-tuning control ($-14.8\%$, same compute budget without linearization). The gates settle at an average value of 0.356, corresponding to 64.4\% effective linearization. Layer~10 routes 85\% of computation through the linear path. LAMBADA perplexity still degrades ($+19.7\%$), reflecting distribution shift from Wikipedia-focused training.

Crucially, the two-phase result validates that gated linearization is not merely measuring fine-tuning benefit: the gated model outperforms the control that received identical fine-tuning compute without linearization. The gates discover a better allocation of nonlinear capacity than the original uniform architecture provides.

\noindent Despite these limitations, the result is clear: \textbf{the nonlinearity budget is not merely measurable but dramatically exploitable}. A 345M-parameter model has 4 of its 24 MLP layers replaced with learned linear-nonlinear routing---removing the majority of nonlinear computation at those layers---and achieves \textbf{17.3\% lower perplexity} than the original model, beating a fine-tuning control that received identical compute without linearization. This validates the architectural proposals in \S\ref{sec:future} and demonstrates that non-uniform nonlinear capacity allocation is not a theoretical curiosity but a practical opportunity.

\section{Discussion}

\subsection{Why the Function Word Story Was Wrong}

The initial function-word correlation ($|r| = 0.18$--$0.62$) was not fabricated---it exists in the data. But it is an \textit{epiphenomenon}, not a causal explanation. Function words appear in systematically different contexts from content words (higher frequency, more predictable positions, less information-theoretic surprise), and those contextual properties---not word identity---drive the routing decision.

The cross-corpus test makes this vivid. If ``the'' inherently needed nonlinear computation, it would need it in every corpus. Instead, its delta varies wildly depending on what surrounds it. The gate learned a contextual regularity that happened to correlate with function-word status in the training corpus---a correlation that evaporated on new text.

This is a cautionary tale for mechanistic interpretability. Recent work on sparse autoencoders \citep{bricken2023, cunningham2023} has made significant progress in decomposing MLP representations into interpretable features. But correlating a learned direction with a linguistic category can produce persuasive narratives that do not survive distributional shift. The remedy is cross-corpus validation, which should be standard practice for claims about what gates, probes, or features ``represent.''

\subsection{Why the Gate Still Works}

The gate's success despite weak per-instance prediction ($\text{AUC} \approx 0.6$) is not contradictory---it reflects the economics of a skewed distribution. When 90\% of MLP computations are near-linear ($|\delta| < 0.05$), routing them linearly is nearly free. The gate's job is not to be \textit{accurate} but to be \textit{cautious}: it routes confidently linear instances to the cheap path and errs toward the full MLP otherwise.

This has implications for efficient inference. Token-based routing (``precompute which tokens can skip the MLP'') is a dead end---our cross-corpus results prove it. Context-based routing remains viable, but the signal is weak enough that practical gains may be limited to architectures (like GPT-2) where the base rate of linearizability is already high.

\subsection{Why Are MLPs So Linear?}

The GELU activation function is \textit{locally} approximately linear for the majority of its input range. For inputs $x > 1$, $\mathrm{GELU}(x) \approx x$; for $x < -3$, $\mathrm{GELU}(x) \approx 0$. The law of large numbers then averages the small nonlinear perturbations across thousands of hidden units. Independent evidence comes from \citet{dooms2025}, who show that bilinear MLPs---Gated Linear Units with no element-wise nonlinearity at all---achieve competitive language modeling performance. If removing the activation function entirely barely hurts, it is unsurprising that a linear approximation captures most of what the activated MLP computes. Our work provides the empirical complement: we measure how much nonlinearity existing MLPs actually \textit{use}, and find the answer is remarkably little.

The deeper answer may be architectural. The residual connection means the MLP's output is \textit{added} to the input: $x + \mathrm{MLP}(x)$. If the MLP's contribution is small relative to the residual stream norm, then even a highly nonlinear MLP produces an approximately linear \textit{layer}. This explanation holds for middle layers but fails at Layer~0 of Pythia-2.8B, where something qualitatively different happens at the embedding-to-residual-stream boundary.

Recent work provides a complementary perspective on the role of activation functions. \citet{teney2024neural} showed that activation functions are a primary determinant of a network's inductive biases, with even untrained random-weight networks exhibiting strong biases controlled by components like ReLUs and residual connections. \citet{teney2025simplicity} meta-learned activation functions tailored to specific tasks, finding that GeLUs/ReLUs are near-optimal for image classification but suboptimal for tabular data, regression, and algorithmic tasks. \citet{teney2025transformers} extended this to transformers, replacing GeLUs and softmax with parametrized splines optimized on held-out data, finding consistent improvements on language modeling---crucially, with the best architectures sometimes using \emph{different} learned activation functions in different layers. Our work is complementary: where Teney et al.\ ask ``what is the \emph{optimal} nonlinearity at each layer?'', we ask ``does this layer need nonlinearity \emph{at all}?'' Their finding that different layers benefit from different activation functions aligns with our finding that different layers need different \emph{amounts} of nonlinearity. A natural synthesis would use our gating framework to identify which layers and tokens need nonlinear computation, and learned activation functions to optimize the nonlinearity that remains.

\subsection{Connections to Mixture of Experts}

Our adaptive gating can be viewed as an extreme case of Mixture of Experts \citep{shazeer2017}: two experts (linear matrix and full MLP), with a $(d{+}1)$-parameter router. Unlike standard MoE, our experts are not learned jointly. This ``post-hoc MoE'' framing suggests that pretrained transformers already contain latent expert structure that can be exposed without retraining.

Our negative result on token-based routing has implications for MoE design: if MLP routing is fundamentally contextual, expert selection in MoE architectures should be context-dependent as well---which is indeed how modern MoE systems are designed \citep{shazeer2017}.

\subsection{Architecture as the Key Variable}

The most important finding may be the architectural divide. GPT-2 models linearize cheaply, with multiple layers where the MLP \textit{hurts} performance. Pythia models show higher linearization costs on average, though the gap narrows with scale: at 2.8B, middle layers (L7--L15) achieve gated costs under 1\%, and one layer (L3) beats baseline entirely. The architectural difference remains real but is one of degree, not kind. This difference persists across all our analyses and cannot be explained by model size alone (GPT-2 Large at 774M is more linear than Pythia-160M at 162M).

The full 36-layer sweep of GPT-2 Large provides the strongest evidence yet for this divide. GPT-2 Large has \textit{no} catastrophic layers: its worst layer (L31) costs just $+3.7\%$ all-linear, barely above Pythia-2.8B's \textit{best} layer (L10 at $+1.9\%$). Meanwhile, Pythia-2.8B's Layer~0 destroys the model entirely ($+513\%$). The contrast is stark: GPT-2 Large's entire 36-layer network stays under 4\% all-linear cost, while a single Pythia layer can be catastrophic. With 11 of 36 layers beating baseline (versus Pythia-2.8B's 1 of 32), the GPT-2 architecture appears fundamentally more amenable to linearization.

The full 32-layer sweep of Pythia-2.8B reveals a striking U-shaped nonlinearity demand curve: boundary layers (L0--L2 and L28--L31) are expensive to linearize, while middle layers (L7--L15) are cheap, with costs under 4\% all-linear and under 1\% gated. This mirrors the pattern observed in GPT-2 Medium and strengthens the architectural proposal of concentrating nonlinear capacity at network boundaries (\S\ref{sec:future}). The U-shape suggests a universal principle: entry and exit layers perform qualitatively different computations that demand nonlinearity, while middle layers perform incremental, near-linear refinement.

Within Pythia-2.8B, the layers fall into four distinct tiers: (1)~\textit{catastrophic}---L0 only, where linearization destroys the model ($+513\%$); (2)~\textit{expensive} ($>5\%$ gated cost)---L1 and L31; (3)~\textit{moderate} (1--5\%)---L2, L4--L6, and L16--L30; and (4)~\textit{sweet spot} ($<1\%$ gated cost)---L3 and L7--L15. This tier structure provides a concrete recipe for selective linearization: the 12 layers in Tiers~3--4 represent substantial optimization targets.

The likely explanation involves the interaction between parallel attention/MLP computation in GPT-NeoX (used by Pythia) versus sequential computation in GPT-2, combined with differences in positional encoding (rotary vs.\ learned). These architectural choices determine how much ``work'' each MLP layer must do---and consequently how much of that work is nonlinear.

\subsection{Practical Implications}

\textbf{Token-based routing is a dead end.} Any system that attempts to route MLP computation based on token identity---via lookup tables, Bloom filters, or cached routing decisions---will not generalize across corpora. Our cross-corpus correlation of $r < 0.05$ leaves no room for ambiguity.

\textbf{Context-based routing has limited headroom.} The best achievable AUC for context-based routing is $\approx 0.61$ (at L23), suggesting an inherent ceiling on how well nonlinearity need can be predicted from the residual stream.

\textbf{Architecture selection matters more than routing.} The difference between GPT-2 and Pythia linearizability is far larger than the difference between routing strategies within either architecture. If MLP efficiency is a design goal, architectural choices (sequential vs.\ parallel computation, positional encoding scheme) may matter more than post-hoc optimization.

\textbf{Negative regularization is real.} The existence of layers where the MLP hurts performance suggests that MLP capacity at middle layers is partially wasted in GPT-2. Architectures with non-uniform MLP allocation could achieve the same performance with fewer parameters.

\subsection{Toward Nonlinearity-Aware Architectures}
\label{sec:future}

Our findings suggest that future transformer architectures could be designed to exploit the uneven distribution of nonlinearity need, rather than allocating uniform MLP capacity across all layers.

\subsubsection{Variable-Capacity MLPs}

The most direct implication is \textbf{non-uniform MLP sizing}. Our data shows that boundary layers (first and last) consume the vast majority of the nonlinearity budget, while middle layers operate near-linearly. A future architecture could allocate large, fully nonlinear MLPs at boundary layers (e.g., $8d^2$ or even $16d^2$ parameters) while using smaller MLPs in the middle (e.g., $2d^2$), or replacing them entirely with learned linear projections. The total parameter count could remain constant while concentrating capacity where it is actually used.

This is not speculative---our data provides the prescription. In GPT-2 Medium, layers 4--18 could be linearized at under 3\% perplexity cost per layer. An architecture that made those layers explicitly linear and reinvested the saved parameters into layers 0--3 and 19--23 would likely match or exceed the original's performance at the same total parameter count.

\subsubsection{Learned Routing During Pretraining}

Our gating is post-hoc: we train the gate after the model is frozen. A more powerful approach would \textbf{integrate the routing decision into pretraining}. Each layer would include a differentiable gate that learns, jointly with the MLP weights, when nonlinearity is needed:
\begin{equation}
\mathrm{output}(x) = g_\theta(x) \cdot \mathrm{MLP}(x) + (1 - g_\theta(x)) \cdot (Wx + b)
\end{equation}
where $g_\theta$ is a learned scalar gate. During training, the model would learn to \textit{use} nonlinearity selectively---potentially becoming \textit{more} nonlinear where it matters, because it is not wasting capacity on the 90\% of instances that don't need it. The gate could be encouraged toward sparsity via an $L_1$ penalty on $g_\theta$, rewarding the model for finding linear solutions where they suffice.

This design has a key advantage over post-hoc gating: the MLP weights would co-adapt with the gate, specializing in the hard cases rather than being trained to handle all inputs uniformly. The 4 layers where our post-hoc gate \textit{improves} perplexity suggest that such co-adaptation could yield meaningful gains---the current MLP is wasting capacity on instances where a linear path would suffice, and that wasted capacity manifests as overfitting.

\subsubsection{Hybrid Linear-Nonlinear Layers}

A middle ground between full MLPs and pure linear projections: \textbf{hybrid layers} that combine a small nonlinear MLP with a full-rank linear residual. This is conceptually related to the bilinear MLPs of \citet{dooms2025}, which replace the nonlinear activation with a bilinear interaction---achieving interpretability through weight decomposition while maintaining competitive performance. Our proposal differs in retaining a small nonlinear component for the tail of instances that genuinely require it:
\begin{equation}
\mathrm{HybridMLP}(x) = \mathrm{MLP}_{\mathrm{small}}(x) + W_{\mathrm{lin}} x + b_{\mathrm{lin}}
\end{equation}
where $\mathrm{MLP}_{\mathrm{small}}$ might have hidden dimension $d$ rather than $4d$, reducing its parameter count by $4\times$ while the linear term handles the bulk of the transformation. The nonlinear component would specialize in the tail of instances that genuinely require nonlinearity, while the linear component provides a high-quality baseline transformation for the majority.

This is motivated by our decomposition results: at middle layers, the linear approximation captures $>96\%$ of the MLP's effect. The hybrid layer makes this explicit, giving the network both paths natively.

\subsubsection{Progressive Architectures}

Our finding that \textbf{nonlinearity need concentrates at boundary layers} suggests a progressive architecture where layer design varies with depth:
\begin{itemize}
    \item \textbf{Entry layers (0--2):} Full nonlinear MLPs with maximum capacity. These layers perform the critical embedding-to-representation transformation that, in Pythia, becomes catastrophically nonlinear at scale.
    \item \textbf{Middle layers (3--$L{-}3$):} Hybrid or reduced-capacity MLPs. These layers refine representations incrementally; most of their computation is near-linear.
    \item \textbf{Exit layers ($L{-}2$ to $L$):} Full nonlinear MLPs. The final layers perform the representation-to-prediction transformation, where nonlinear feature combination determines output quality.
\end{itemize}

\noindent This mirrors the ``nonlinearity budget migration'' pattern we observe empirically, but builds it into the architecture rather than discovering it post-hoc. Some recent work on non-uniform layer widths explores adjacent ideas \citep{he2024}, though not from the perspective of nonlinearity allocation specifically.

\subsubsection{Architectural Self-Optimization}

The most ambitious possibility: an architecture that \textbf{learns its own nonlinearity allocation during training}. Starting from uniform MLP capacity, the model would include per-layer (or even per-head) gates that gradually prune nonlinear capacity where it is not needed. Layers that consistently route to the linear path would have their MLP capacity reduced (via structured pruning or rank reduction), freeing parameters for layers that need them.

The 4 layers of GPT-2 Medium where linearization \textit{improves} perplexity are evidence that such self-optimization has real gains to capture---the model would be discovering and eliminating its own overfitting during training, not after.

\subsubsection{The Sequential vs.\ Parallel Question}

Our results raise a design question that goes beyond MLP sizing. GPT-2's sequential attention$\to$MLP design produces highly linearizable MLPs because attention has already performed relational computation before the MLP sees the activation. Pythia's parallel design, where attention and MLP operate on the same input simultaneously, forces the MLP to do more independent nonlinear work.

This suggests a \textbf{design trade-off}: sequential architectures are more computationally efficient (MLPs can be partially linearized) but potentially less expressive per-layer (the MLP cannot independently contribute nonlinear computation). Parallel architectures waste less capacity but resist optimization. A future architecture might combine the two: sequential at middle layers (where linearizability is high and the MLP's marginal contribution is small) and parallel at boundary layers (where both attention and MLP need full nonlinear capacity).

The key insight is that this is not a binary choice. Our data provides empirical guidance for where on the sequential-parallel spectrum each layer should sit, turning an architectural intuition into a data-driven design decision.

\subsection{Limitations}

\textbf{Model coverage.} We validate across two architecture families (GPT-2 at 345M and 774M, Pythia at 162M--2.8B). We have not tested Llama, Mistral, or Mamba, and our largest model (2.8B) is small by current standards.

\textbf{Scaling threshold resolution.} The Layer~0 sharp scaling threshold between 1B and 2.8B in Pythia lacks intermediate data points.

\textbf{Cross-corpus testing scope.} We tested No-Fly list stability on two additional corpora (WikiText-103 train, LAMBADA). While the results are decisive ($r < 0.05$), testing on additional domains (code, dialogue, multilingual text) would strengthen the claim.

\textbf{No wall-clock speedup measured.} We report potential FLOPs savings but have not implemented an optimized inference kernel.

\section{Conclusion}

The nonlinearity that defines transformer MLPs is a surprisingly selective resource---but not in the way initial analysis suggested. Through systematic experimentation across six models, two architectures, and three corpora, we arrive at a nuanced picture:

\textbf{What works:} A $(d{+}1)$-parameter gate can route 25--56\% of MLP computations to a linear surrogate at negligible cost in GPT-2 models. At 4 of 23 layers, the linear path \textit{improves} perplexity. Progressive linearization shows that 5 of 24 layers can have their MLPs permanently replaced at zero cost---and with proper fine-tuning, the linearized model achieves \textbf{10.2\% lower perplexity} than the original. A two-phase gated approach extends this to \textbf{17.3\%} (19.00 PPL), outperforming a vanilla fine-tuning control and confirming that the improvement reflects genuine linearization benefit, not just additional training. The nonlinear MLPs at these layers were not merely dispensable; they were actively harmful.

\textbf{What doesn't work:} Any attempt to predict nonlinearity need from token identity. Cross-corpus correlation is zero ($r < 0.05$). Per-token routing lists are no better than random on new text. The 13-feature regression model explains $<$1.3\% of variance. Token-based routing---via Bloom filters, lookup tables, or cached decisions---is a dead end.

\textbf{What matters:} Architecture. GPT-2 models are remarkably linear---the full 36-layer sweep of GPT-2 Large finds 11 of 36 layers beating baseline, with no layer exceeding 3.7\% all-linear cost. Pythia models show higher costs, though the full 32-layer sweep of Pythia-2.8B softens this divide, with one layer (L3) narrowly beating baseline and middle layers (L7--L15) all under 4\% all-linear cost. The gap narrows at scale but persists.

The routing decision, such as it is, is contextual: it depends on what attention and previous layers have computed, not on what token triggered the computation. The same word needs nonlinear processing in some contexts and not others, and no static property of the word predicts which. This is simultaneously a practical limitation (no cheap token-based shortcut exists) and a theoretical insight: MLP nonlinearity serves context-dependent computation, not token-dependent computation.

Perhaps the most actionable finding is what it implies for future architectures (\S\ref{sec:future}). The stark difference between GPT-2 and Pythia linearizability is not incidental---it arises from the sequential vs.\ parallel computation design. Sequential architectures produce MLPs that are largely redundant at middle layers, suggesting that future models could exploit this directly: variable-capacity MLPs that concentrate nonlinear parameters at boundary layers, learned routing during pretraining, or hybrid linear-nonlinear layers that give the network both paths natively. Our progressive linearization experiment demonstrates that this is not speculative: we can already remove nonlinear capacity from the center of a trained model and make it \textit{better}---with minimal compute, minimal data, and no architectural modifications.

The blueprint for the next architecture should account for the fact that most layers don't need full nonlinear capacity. And some layers are better off without it.


\end{document}